# How Does Traffic Environment Quantitatively Affect the Autonomous Driving Prediction?

Wenbo Shao, Yanchao Xu, Jun Li, Chen Lv, *Senior Member, IEEE*, Weida Wang and Hong Wang⊠, *Senior Member, IEEE*

***Abstract*—An accurate trajectory prediction is crucial for safe and efficient autonomous driving in complex traffic environments. In recent years, artificial intelligence has shown strong capabilities in improving prediction accuracy. However, its characteristics of inexplicability and uncertainty make it challenging to determine the traffic environmental effect on prediction explicitly, posing significant challenges to safety-critical decision-making. To address these challenges, this study proposes a trajectory prediction framework with the epistemic uncertainty estimation ability that outputs high uncertainty when confronting unforeseeable or unknown scenarios. The proposed framework is used to analyze the environmental effect on the prediction algorithm performance. In the analysis, the traffic environment is considered in terms of scenario features and shifts, respectively, where features are divided into kinematic features of a target agent, features of its surrounding traffic participants, and other features. In addition, feature correlation and importance analyses are performed to study the above features' influence on the prediction error and epistemic uncertainty. Further, a cross-dataset case study is conducted using multiple intersection datasets to investigate the impact of unavoidable distributional shifts in the real world on trajectory prediction. The results indicate that the deep ensemble-based method has advantages in improving prediction robustness and estimating epistemic uncertainty. The consistent conclusions are obtained by the feature correlation and importance analyses, including the conclusion that kinematic features of the target agent have relatively strong effects on the prediction error and epistemic uncertainty. Furthermore, the prediction failure caused by distributional shifts and the potential of the deep ensemble-based method are analyzed.**

## I. Introduction

TRAJECTORY prediction is an indispensable part of the autonomous driving pipeline [1]. To drive safely and efficiently in complex traffic environments, autonomous vehicles (AVs) are required to have the ability to predict the future motion of surrounding traffic participants (TPs), such as vehicles and pedestrians, accurately and reliably. In recent years, with the accumulation of large-scale driving data and rapid development of algorithms, artificial intelligence (AI) has been widely applied to autonomous driving trajectory prediction [2, 3], and promising results have been achieved. However, trajectory prediction has still been challenging, particularly in urban driving scenarios, where an agent's movement is influenced by a combination of its historical state and its complex interactions with the surrounding environment. Many recent studies [4-6] have considered multiple factors simultaneously to improve trajectory prediction algorithms, but there has still been certain performance degradation of a prediction model in complex traffic environments.

With the improvement in prediction accuracy, the complexity of AI-based models has also increased gradually. Highly elaborated models pose a great challenge to explaining the operation and failure mechanisms of prediction algorithms, which in turn reduces the credibility of a prediction model. In addition, AI has its inherent uncertainty and faces many problems, such as insufficient training data, imperfect model architecture, and limited training process, which may lead to functional insufficiencies of the model under specific environmental conditions, potentially causing severe traffic accidents [7]. The existing research mainly focuses on improving the dataset-level accuracy of prediction algorithms [8, 9], and little attention has been paid to the changes in prediction performance under different environmental conditions. However, this is not conducive to addressing the practical challenges that a prediction algorithm confronts.

For a target agent (TA) moving in a specific scenario, a trajectory prediction model predicts its future trajectory by modeling time series, interaction, and other relationships based on its historical state, surrounding TPs' features, and other environmental features. Correspondingly, various traffic environmental factors may have different effects on trajectory prediction, but fewer studies have quantitatively investigated these effects. Further, data-driven methods strongly depend on

This work has been submitted to the IEEE for possible publication. Copyright may be transferred without notice. This work was supported in part by the National Science Foundation of China Project: 52072215and U1964203, and the National Key R&D Program of China:2020YFB1600303. (*Corresponding authors: Hong Wang*)

Wenbo Shao, Jun Li and Hong Wang are with Tsinghua Intelligent Vehicle Design and Safety Research Institute, School of Vehicle and Mobility, Tsinghua University, Beijing 100084, China. (e-mail: swb19@mails.tsinghua.edu.cn; lijun1958@tsinghua.edu.cn; hong_wang@tsinghua.edu.cn).

Yanchao Xu and Weida Wang is with the School of Mechanical Engineering, Beijing Institute of Technology, Beijing 100081, China. (e-mail: 3120200410@bit.edu.cn, wangwd0430@163.com)

Chen Lv is with the School of Mechanical and Aerospace Engineering, Nanyang Technological University, Singapore 639798 (e-mail: lyuchen@ntu.edu.sg).

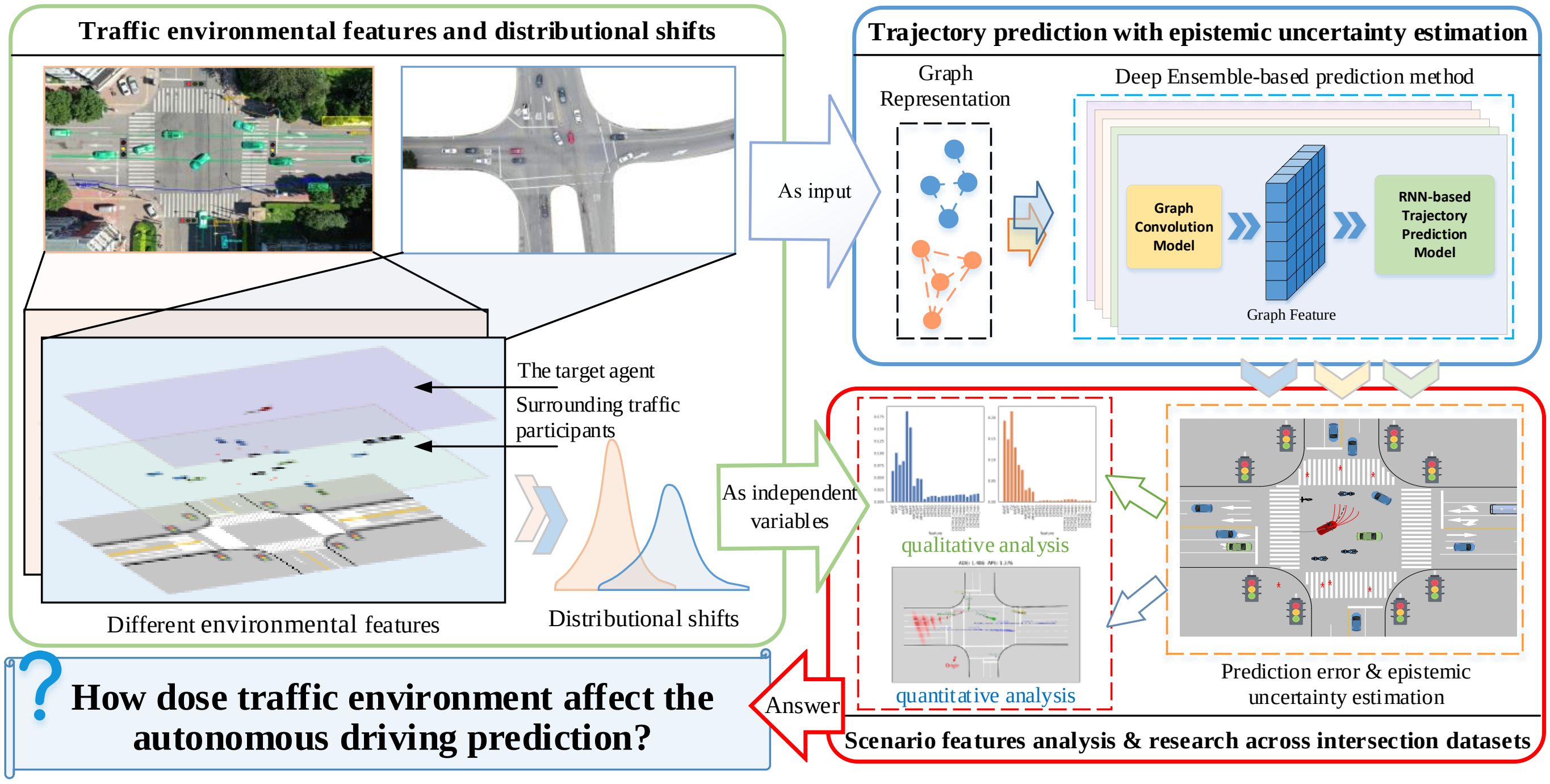

**Fig. 1.** Illustration of the traffic environment effect on the trajectory prediction algorithm performance. The traffic environmental data include various TAs' states, their surrounding TPs' states, and other contextual information, which may affect the prediction differently. In addition, variations in time and place can lead to distributional shifts, which may further degrade the prediction performance. This study focuses on extracting these factors and analyzing their influence on prediction performance.

training data, and a model trained on one dataset may not perform well on other datasets. In real-world applications, the operating environment of AVs may change significantly with different factors, such as time, geography, country, and weather conditions. This may cause certain distributional shifts, posing additional challenges to trajectory prediction. Therefore, it is increasingly important to study how distributional shifts [10] in a real environment affect trajectory prediction. As shown in Fig. 1, this work focuses on the effects of both specific scenario features and scenario shifts on the prediction algorithm.

As for the prediction algorithm performance, previous studies have generally focused on prediction error. In recent years, there has been an increasing interest in extracting the uncertainty of AI-based models [11, 12], thus empowering the models with a self-awareness ability. Epistemic uncertainty [13] is a recurring suggestion that helps to represent the model's confidence in its current predictions; namely, these models tend to have greater epistemic uncertainty when they encounter challenging environments. Therefore, the epistemic uncertainty of a prediction model is extracted and considered a type of performance metric in this work. As shown in Fig. 1, based on both the prediction error and epistemic uncertainty, the effect of the traffic environment on the prediction algorithms' performances can be analyzed.

The main contributions of this work can be summarized as follows:

- A trajectory prediction framework that integrates epistemic uncertainty estimation is proposed. The proposed framework performs the TA's future state prediction and estimates the epistemic uncertainty simultaneously;
- The potential of the proposed deep ensemble-based trajectory prediction framework for improving the prediction algorithm robustness and estimating epistemic uncertainty is demonstrated;
- For the trajectory prediction task, the key features of a traffic environment are extracted, and methods for feature correlation analysis and feature importance analysis are proposed to obtain the relationship between the traffic environment and the trajectory prediction algorithm performance;
- The distributional shifts between different intersection datasets and the resulting trajectory prediction degradation are investigated. The features of multiple datasets and their prediction difficulty levels are analyzed, and it is demonstrated that the deep ensemble is helpful in improving the trajectory prediction robustness against cross-dataset evaluation.

The remainder of this paper is organized as follows. Section II presents the existing work related to this paper. Section III introduces the proposed method. Section IV describes the datasets and evaluation metrics used in this work, as well as implementation details. Section V analyzes and discusses the experimental results. Section VI concludes the paper.

## II. Related Work

### *A. Trajectory Prediction*

There have been numerous studies on improving the trajectory prediction algorithms, and according to the modeling principles, they can be mainly divided into physics-based methods, maneuver-based methods, and interaction-aware methods [14]. Physics-based methods [15] consider only the historical motion state of an object while ignoring the influence of surrounding TPs. Therefore, they are mainly suitable for short-term trajectory predictions. Maneuver-based methods [16] learn prototype trajectories from the observed agent behaviors to predict future motion, but they lack consideration of interactions between TPs. Interaction-aware methods [17] have shown better performance compared to the other two types of methods through learning the interaction between a TA and surrounding TPs.

In recent works, many methods have been used to model interactions between agents, providing valuable information for trajectory prediction improvement [3, 9]. For instance, social pooling (S-pooling) [8] pools hidden states of a TA's neighbors within a certain spatial distance to model interactions with the surrounding environment. Convolutional social pooling [18] combines the convolutional and max-pooling layers to model interactions between agents in the occupancy grid. Subsequently, the grid representation is further modified to consider only eight neighbors that have the most critical impact on the TA [19]. In addition, recent research has focused on the rasterized representation of scenes; the historical state of a TA and scene context were co-encoded in a raster map [20-22], and various information was distinguished by different channels and colors. In addition, convolutional neural networks (CNNs) were used to extract desired features from raster graphs. Graph models have attracted great interest recently due to their good performance in modeling inter-agent interactions. In graph models, a node represents an agent, and an edge represents an interaction between two agents. Diehl *et al.* [23] modeled interactions between vehicles as a homogeneous directed graph to achieve high computational efficiency and large model capacity. They evaluated graph convolutional networks and graph attention networks and introduced several adaptations for specific scenarios. Mo *et al.* [2] employed a heterogeneous edge-enhanced graph attention network to handle the heterogeneity of TAs and TPs. The GRIP [24] represents the input as a specific grid and uses an undirected graph to model interactions between agents within a certain range, where fixed graphs are considered in the graph convolution submodule. The GRIP++ [5] improves the above-mentioned method by adopting trainable graphs, which overcomes the shortcoming that fixed graphs based on manually designed rules cannot model interaction properly. In addition to the interaction modeling, another important requirement of trajectory prediction relates to time series processing. Recently, recurrent neural networks (RNNs), including the long short-term memory (LSTM) and gated recurrent unit (GRU) models, have been widely used in modeling sequential problems, and significant results have been achieved. Accordingly, these models have been used as sub-modules in many trajectory prediction algorithms [5, 6].

Neural networks have been demonstrated to be highly efficient in trajectory prediction for different classes of TPs. Research on pedestrian intent modeling and motion prediction has been conducted for decades. The Social-LSTM [8] is a typical success case in early research in this field, which combines S-pooling and LSTM to predict the future trajectory of pedestrians in crowded scenes. The Social-GAN [25] uses generative adversarial networks (GANs), sequence-to-sequence models, and pooling mechanisms and employs the corresponding generators and recursive discriminators to predict pedestrians' socially feasible future. However, the GAN model training is difficult and may not converge and can lead to mode collapsing and dropping. Therefore, the Social-Ways uses the Info-GAN, which does not apply the mean square error loss (L2 loss) to force the generated samples to be close to real data but adds another item to consider mutual information, thus alleviating the above-mentioned problems. Since vehicles have higher running speeds and need to obey more road constraints than pedestrians, predicting their future movements is a prerequisite for realizing safe and efficient autonomous driving. A number of studies have designed specialized networks for vehicle trajectory prediction [26]. For instance, vehicle trajectory prediction in highway scenarios, which are relatively simple and where the motion pattern of a vehicle is relatively fixed, has received early attention [18, 24, 27]. With the collection of large-scale datasets [28, 29] and the development of autonomous driving in urban scenes, much research has been focused on motion prediction in complex urban environments [4, 30-32]. TrafficPredict [33] adopted a four-dimensional graph to model the interaction in the instance and category layers, thus realizing the heterogeneous traffic-agent trajectory prediction. The GRIP++ achieved joint trajectory prediction of all observed objects while considering multiple classes of TPs, thus greatly improving real-time prediction performance. However, the above work focuses on the improvement in the dataset-level accuracy while ignoring the research on the sensitivity of the prediction algorithm to environmental factors, which is the focus of this work.

### *B. Epistemic Uncertainty Estimation*

Due to the rapid development of neural networks and their application to trajectory prediction tasks, it has become increasingly important to estimate the network confidence in its prediction accuracy. However, the original neural network cannot provide an estimation of its epistemic uncertainty. To address this shortcoming, some studies have considered and quantified the epistemic uncertainty of neural networks [11, 34, 35], which represents an indicator that can express how confident the network is in its current prediction result. The main epistemic uncertainty estimation methods include the Bayesian neural network (BNN), single-pass uncertainty estimation, and ensemble-based methods. The BNN quantifies the epistemic uncertainty of a neural network by introducing uncertainty into its parameters. The key challenge of these methods is to solve the posterior distribution of network parameters. In the early research, variational inference (VI) [36], which uses a prespecified family of distributions [37, 38], was widely adopted as a method with a strong theoretical basis. However, with the rapid growth in the neural network structure complexity, VI has faced many challenges in terms of solving difficulty and computational complexity. To address these limitations, the Monte Carlo (MC) dropout [39, 40] was proposed to approximate the results obtained by sampling, assuming that the network weights conformed to a Bernoulli distribution. It has been theoretically demonstrated that the MC dropout has the ability to approximate epistemic uncertainty. In single-pass uncertainty estimation, uncertainty is obtained through one forward propagation, which has obvious advantages in terms of computational complexity. The deep evidence theory is a representative method and has been widely used in classification [41] and regression [42] tasks. However, these methods require that the original network output has a specific form, which limits their scalability. In addition, these methods do not consider the uncertainty of network weights. In view of that, some studies [43] positioned the uncertainty they extracted as distributional uncertainty, different from epistemic uncertainty. In deep ensemble-based methods, the training process is adjusted to obtain multiple different models, and epistemic uncertainty is estimated by synthesizing the prediction results of the models. Deep ensemble [44] is a simple and scalable uncertainty estimation method, which has attracted extensive attention due to its excellent performance in estimating epistemic uncertainty [45]. Currently, this method has become a mainstream paradigm. Subsequently, to reduce the storage and computational costs of the practical application of deep ensemble, many improved methods have been proposed [46, 47]. For instance, the Batch-Ensemble [46] reduces training and testing costs by defining each weight matrix as the Hadamard product of the shared weights of all ensemble members and the rank-one matrix of each member, but the uncertainty estimation performance is slightly decreased. However, previous studies on epistemic uncertainty have usually involved tasks such as semantic segmentation and object detection but have lacked detailed research in the field of trajectory prediction. This work proposes a trajectory prediction method with epistemic uncertainty estimation, where deep ensemble and MC dropout are used separately to estimate epistemic uncertainty and compared on the real intersection dataset.

### *C. Relationship between Prediction* Performance *and Traffic Environment*

Previous studies have mainly focused on enhancing the dataset-level accuracy of trajectory prediction. However, the actual trajectory prediction performance can be strongly dependent on a traffic environment. Therefore, it is of great significance to determine the relationship between the environment and prediction model to improve the interpretability of trajectory prediction algorithms and determine their limitations. This is essential for safety-critical autonomous driving applications. Several works focused on modeling and complexity calculation of a traffic environment using different methods, such as five- and six-layer scene models [48, 49], where layer elements can have a strong correlation with the prediction algorithm. Wang *et al.* [50] proposed a method to quantify scenario complexity in traffic but did not explore its relationship with the autonomous driving algorithm performance. The Shapley value is a feature attribution method that helps to measure the contribution of input variables to model performance. Makansi *et al.* [51] proposed a variant of Shapley value and analyzed the problems that some of the existing trajectory prediction models consider only the past trajectory of a TA and are difficult to reason about interactions. In addition, recent studies have gradually paid attention to the cross-dataset performance of AI algorithms in object detection and prediction applications [52, 53]. Gesnouin *et al.* [54] evaluated the impact of differences in pedestrian poses and detection box heights in different datasets on pedestrian crossing prediction. Gilles *et al.* [10] compared the accuracy of vehicle trajectory prediction algorithms on several datasets containing mixed scenarios. However, there has still been a lack of comprehensive analysis of traffic environmental factors and their changes and quantitative research on their impact on prediction algorithms.

n this work, the research scenario is the intersection, which is a typical and challenging urban scenario. Distributional shifts between different intersection datasets and their effect on trajectory prediction performance, considering both error and epistemic uncertainty, are analyzed.

## III. Proposed Method

### *A. Trajectory Prediction with Epistemic Uncertainty Estimation*

#### *1) Trajectory Prediction*

Trajectory prediction is a task that estimates a TA's future position based on historical data on its state and context in a scenario. Particularly, at time $t=0$, the historical input state $S$ of a TA (over previous $t_h$ time steps) is represented as follows:

$$\mathbf{S}=\left[s^{(-t_h+1)},s^{(t_h+2)},\cdots,s^{(0)}\right], \tag{1}$$

where $s^{(t)}$ is the state of a TA at $t$, and it is defined as $s^{(t)}=\left[x^{(t)},y^{(t)}\right]$.

In addition, interactions between a TA and its surrounding environment are modeled based on scene context information $\mathbf{C}$, which includes information on the states of TPs' around the TA and environmental conditions.

A trajectory prediction model $f$ is trained on dataset $\mathcal{D}$. Based on the input $\mathbf{X}=\left[\mathbf{S},\mathbf{C}\right]$, the trained prediction model outputs an estimate $\hat{\mathbf{Y}}$ of the real future trajectory $\mathbf{Y}$ of the TA as follows:

$$\hat{\mathbf{Y}}=f\left(\mathbf{X}\right)=f\left(\mathbf{X},\mathcal{D}\right)=f\left(\mathbf{X},\hat{\theta}\right), \tag{2}$$

where $\hat{\mathbf{Y}}=[\hat{s}^{(1)},\hat{s}^{(2)},\cdots,\hat{s}^{(t_f)}]$, $t_f$ is the predicted horizon, and $\hat{\theta}$ represents the trained model parameters.

In this work, the GRIP++, which is an enhanced graph-based interaction-aware trajectory prediction method, is used as a base model. It uses both fixed and dynamic graphs to describe the relationship between different TPs, considering the effect of inter-agent interactions on a TA's motion. Furthermore, this method employs a GRU-based encoder-decoder architecture as a sub-module and allows joint trajectory predictions for multiple agents, achieving good performances in terms of prediction speed and accuracy.

#### *2) Epistemic Uncertainty Estimation*

In the previous section, a neural network-based trajectory prediction model is presented, but the original GRIP++ can output only deterministic prediction results. However, real-world traffic scenarios are complex and variable, and it is difficult to construct a training set that will effectively cover all scenarios. In addition, deep learning-based models are inherently uncertain and difficult to interpret, so they may not be reliable enough when confronted with unknown scenarios (e.g., scenarios unseen during training or scenarios with only limited available information). These problems can result in unacceptable degradation in autonomous driving performance. In this regard, the BNN models and learns the posterior distribution of network weights $\hat{\theta} \sim P(\theta \mid \mathcal{D})$, which can be used to estimate the epistemic uncertainty as follows:

$$\hat{\mathbf{Y}} = f(\mathbf{X}, \mathcal{D}) = \int f(\mathbf{Y} \mid X, \theta) \mathrm{P}(\theta \mid \mathcal{D}), \quad (3)$$

where the main issue is how to learn the posterior distribution of parameters effectively.

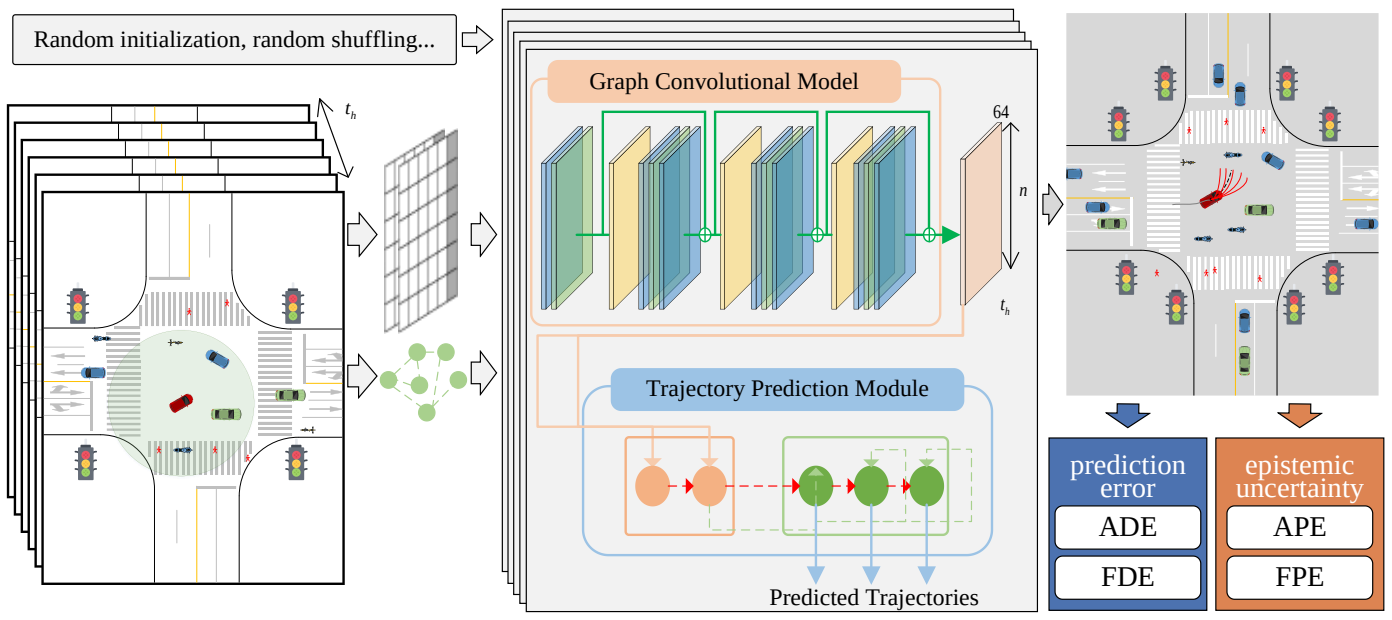

**Fig. 2.** The trajectory prediction framework with epistemic uncertainty estimation (deep ensemble-based method).

The Bayesian approximate inference is a typical solution, which learns an approximate distribution $q(\theta)$ of $\mathrm{P}(\theta \mid \mathcal{D})$. The MC dropout has been shown to be an effective sample-based method for approximate inference, where the network weights are assumed to follow the Bernoulli distribution. After adding appropriate regularization during training and turning on dropout during testing, epistemic uncertainty can be estimated by sampling multiple times.

In recent years, deep ensemble, as a simple, parallelizable, and scalable method, has shown excellent uncertainty estimation ability. In this work, a deep ensemble-based uncertainty estimation framework for the trajectory prediction model is proposed. Specifically, random initialization of neural network parameters and random shuffling of a dataset are performed because they have been proven to have enough good performance in practice. After training, $K$ models of isomorphism and different parameters are obtained. Further, by integrating the results of $K$ models, the final trajectory prediction output is obtained by,

$$\hat{\mathbf{Y}} = K^{-1} \sum_{k=1}^{K} f\left(\mathbf{Y} \mid \mathbf{X}, \hat{\theta}_k\right), \quad (4)$$

where $\hat{\theta}_k$ denotes the post-training parameters of the $k$th model among the ensemble models; similarly, in the MC dropout-based method, $\hat{\theta}_k$ indicates the model parameters for the $k$th dropout during testing.

The predictive entropy is employed to quantify the epistemic uncertainty of the proposed prediction model, where entropy increases with uncertainty. The proposed model outputs $K$ continuous trajectories $\hat{\mathbf{Y}}_k = [\hat{s}_k^{(1)}, \hat{s}_k^{(2)}, \cdots, \hat{s}_k^{(t_f)}]$ in a prediction task, each of which contains the predicted position at multiple future moments. To realize a prediction-task-wise uncertainty estimation, the predictive entropy at multiple moments is integrated to obtain the average predictive entropy (APE) as follows:

$$\mathrm{APE} = \frac{1}{t_f} \sum_{i=1}^{t_f} \mathcal{H}\left[\hat{s}^{(t)}\right] = \frac{1}{t_f} \sum_{i=1}^{t_f} -\int p\left(\hat{s}^{(t)}\right) \ln p\left(\hat{s}^{(t)}\right) \mathrm{d}\hat{s}_t. \quad (5)$$

Assuming that $\hat{s}^{(t)} = \left[\hat{x}^{(t)}, \hat{y}^{(t)}\right]$ obeys the two-dimensional Gaussian distribution, $\hat{x}^{(t)}$ and $\hat{y}^{(t)}$ are independent of each other, and the APE can be expressed as follows:

$$\begin{aligned} \mathrm{APE} &= \frac{1}{t_f} \sum_{i=1}^{t_f} (\ln 2\pi + 1) + \frac{1}{2} \ln \left|\hat{\Sigma}^{(t)}\right| \\ &= \frac{1}{t_f} \sum_{i=1}^{t_f} (\ln 2\pi + 1) + \frac{1}{2} \ln \sigma^2\left(\hat{x}^{(t)}\right) \sigma^2\left(\hat{x}^{(t)}\right). \end{aligned} \quad (6)$$

Similarly, the final predictive entropy (FPE) is defined as,

$$\begin{aligned} \mathrm{FPE} &= \mathcal{H}\left[\hat{s}^{(t_f)}\right] = (\ln 2\pi + 1) + \frac{1}{2} \ln \left|\hat{\Sigma}^{(t_f)}\right| \\ &= (\ln 2\pi + 1) + \frac{1}{2} \ln \sigma^2\left(\hat{x}^{(t_f)}\right) \sigma^2\left(\hat{x}^{(t_f)}\right). \end{aligned} \quad (7)$$

### *B. Scenario Features Extraction*

In prediction scenarios, a TA's motion is related to its historical state, interactions with surrounding TPs, and other factors. Although the existing prediction algorithms have either explicitly or implicitly considered different factors, their performances may still be affected by the above-mentioned features due to algorithm limitations. Therefore, three types of scenario features are considered in this work: 1) dynamic features of a TA, which include data on its historical or future motion states; 2) features of surrounding TPs, which refer to their states and interactions with the TA; 3) other scenario features, which include the type, behavior pattern, compliance with traffic rules, and current location of the TA.

#### *1) Kinematic features of TA*

The historical data on the TA motion state denote an important input to a prediction model and have a direct impact on the model output. In addition, the future motion state of a TA is a key reference for evaluating the model's prediction accuracy. Therefore, in this work, the kinematic features of TA are extracted to analyze their impact on the prediction algorithm performance.

Velocity is one of the primary kinematic features, which directly reflects the aggressiveness of TA movement. It also represents the discrete degree of a continuous trajectory. Considering the trajectory prediction model characteristics, three velocity sub-features are extracted: 1) average velocity of the historical trajectory (AVHT), which indicates the aggressiveness of the model's input trajectory; 2) current velocity (CV), which directly represents a TA's current state and has a key impact on the trajectory prediction output; 3) average velocity of the future trajectory (AVFT), which reflects the spatial span of the future trajectory points.

In addition, the velocity variations indicate trajectory stationarity, which may have a significant influence on prediction results. For instance, a sudden start of a parked vehicle may be difficult for the model to predict timely and accurately. Therefore, the acceleration value at each moment is calculated to obtain the following sub-features: 1) average acceleration of the historical trajectory (AAHT), which represents the speed mutation degree of the input trajectory; 2) average acceleration of the future trajectory (AAFT), which reflects the overall situation of a TA's future speed mutation; 3) maximum acceleration of the future trajectory (MAFT), considering that a sudden speed change at any moment can lead to severe deformation of the overall trajectory, it is necessary to extract the fastest speed change in the future as a feature for analysis.

Similarly, changes in the TA moving direction denote a potentially influential factor of prediction performance. For instance, a vehicle going straight may suddenly swerve or make a U-turn, thus posing a serious challenge to the prediction algorithm. Therefore, the heading change speed (HCS) is extracted for analysis. In detail, the absolute value of the change speed of the heading angle at each moment is calculated and used as a basic feature, and then the analysis of the following parameters is performed: 1) average HCS of the historical trajectory (AHCSHT); 2) average HCS of the future trajectory (AHCSFT), which reflects the overall curvature or volatility of the future trajectory; 3) maximum HCS of the future trajectory (MHCSFT), which increases when there is a sudden large change in direction at any point in the future.

*2) Features of Surrounding TPs*

Convoluted interactions with other agents increase the difficulty in trajectory prediction, and although many of the existing prediction methods can explicitly or implicitly model interactions, it has not been fully discussed whether the performance of these black-box models is sensitive to actual interactions. To examine this situation, a set of hierarchical prediction scenario complexity metrics is proposed to analyze the effect of a TA's interactions with surrounding agents on the prediction algorithm performance.

First, with a TA as a center, the prediction scenario complexity has a positive correlation with the number of its surrounding TPs, and a basic feature, the number of TPs within $x$ meters from the TA (NTPx), is defined.

In addition, the distance between the TA and its surrounding TPs directly affects the prediction scenario complexity. Assuming that a set of TPs within $x$ meters of a TA $i$ is denoted by $N_x(i)$, then for any $j \in N_x(i)$, its distance from $i$ is $d_j = dist(s_i, s_j)$. The density of TPs within $x$ meters around TA (DTPx) is given by,

$$\mathrm{DTPx} = \sum_{j=1}^{N_x(i)} e^{-\lambda d_j}, \tag{8}$$

where $\lambda$ is a scaling factor.

Furthermore, the potential conflicts due to the movement of surrounding TPs are analyzed. As shown in Fig. 3, for a TP $j \in N_x(i)$ within $x$ meters around a TA $i$, its current state is given by $[s_j^{(0)}, v_j^{(0)}]$; then, its position after $t$ seconds is expressed as $s_j^{(t)} + v_j^{(t)} t$, and the degree of conflict from TPs within $x$ meters around TA (DCTPx) is defined as follows,

$$\mathrm{DCTPx} = \sum_{j=1}^{N_X(i)} e^{-\lambda\, \mathrm{agg}^{(\mathrm{T})}\left(\alpha^{(t)} d_j^{(t)}\right)}, \tag{9}$$

where $d_j^{(t)} = dist\left(s_i^{(t)}, s_j^{(t)}\right)$; $T$ is the time horizon used for evaluation; $\alpha^{(t)}$ is the scaling factor for the distance at time $t$, and in this study, it is set to grow faster over time to reinforce the focus on the short-term risk; $\mathrm{agg}^{(\mathrm{T})}$ represents the aggregation operation, which is used to synthesize the conflicts at $T$ times in the future. The two basic modes used in this study include the mean value (DCTPx_mean) and the maximum value (DCTPx_max).

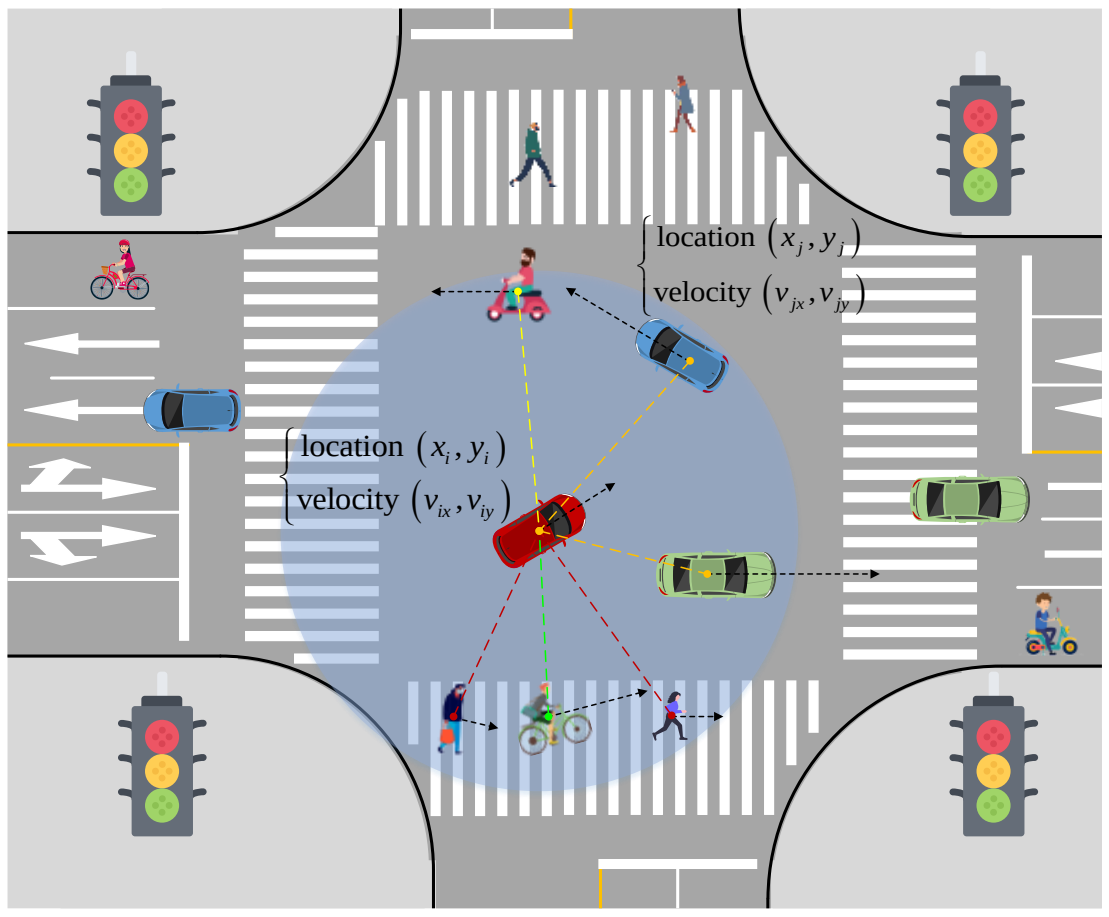

**Fig. 3.** Schematic diagram of the conflict degree calculation from TPs within $x$ meters from a TA.

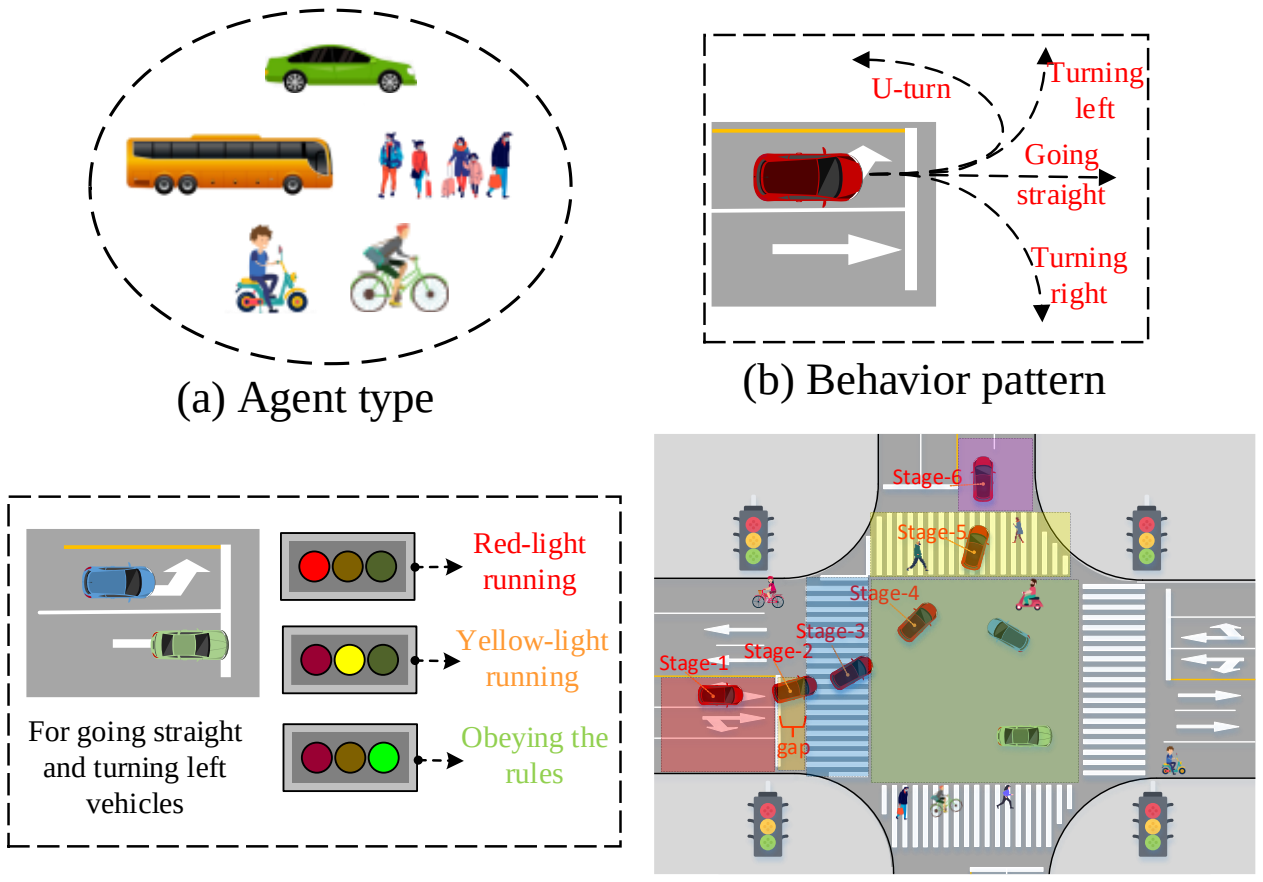

(a) Agent type (b) Behavior pattern

(c) Compliance with traffic rules (d) Agent location

**Fig. 4.** Illustration of the other extracted scenario features.

*3) Other Scenario Features*

In addition to the above two categories of features, the impact of several other representative features on the prediction error and epistemic uncertainty is studied, as shown in Fig. 4.

The considered features include:

- TA type: The GRIP++ can simultaneously predict future trajectories of multiple types of TAs, such as vehicles, pedestrians, and cyclists. Each type of agent has its movement pattern, which may cause different prediction performances. Referring to the research presented in [33], TAs can be divided into four types: small vehicles, vehicles, pedestrians, motorcyclists, and bicyclists;
- TA behavior pattern: This study mainly focused on three basic behavior patterns of vehicles at intersections: going straight, turning left, and turning right. In addition, this study extracts certain corner cases, such as U-turns. The prediction errors and epistemic uncertainty under different behavioral patterns are compared and analyzed in a unified manner;

- TA's compliance with traffic rules: Traffic rules partially constrain the behaviors of participants, but in real-world scenarios, some of TAs may violate the rules, thus affecting the trajectory prediction performance. For instance, in a signalized intersection, the behaviors of TAs can be classified based on their compliance with the traffic signal into obeying the rules, yellow-light running, and red-light running;
- TA location: The whole process of a vehicle passing through the intersection is divided according to the time sequence into six stages: stage 1: ex-entering an intersection; stage 2: in the gap; stage 3: in the first crosswalk; stage 4: inside an intersection; stage 5: in the last crosswalk; stage 6: exiting an intersection.

*C. Scenario Features Analysis*

To analyze the relationship between the above-mentioned features systematically, the qualitative and quantitative analysis methods are adopted. The main methods include feature correlation analysis and feature importance analysis based on random forest regression.

*1) Feature Correlation Analysis*

Correlation analysis is to calculate the degree of correlation between two or more feature variables using correlation coefficients as quantitative indicators. Typical correlation coefficients include the Pearson correlation coefficient and Spearman rank correlation coefficient. The Pearson correlation coefficient requires evaluated variables to conform to the normal distribution, but this is a strong assumption that experimental results can hardly satisfy. In contrast, the Spearman rank correlation coefficient does not have such strict requirements on data as the Pearson correlation. Namely, it requires only that observed values of the two variables are paired rank data or rank data transformed from continuous variable observation data. The Spearman rank coefficient can be mainly used in the monotonic relationship evaluation. Specifically, it is assumed that the original data ( $x_i$ , $y_i$ ) are arranged in ascending order, and $R(x_i)$ and $R(y_i)$ are defined as the ranking of $x_i$ and $y_i$ in their corresponding data, respectively; $\overline{R(x)}$ and $\overline{R(y)}$ denote the mean of the ranking of the two groups of data, and $n$ is the number of data pairs. Then, the Spearman rank correlation coefficient $\rho$ can be defined as follows:

$$\begin{aligned}\rho &= \frac{\sum_{i=1}^{n}\left(R(x_i)-\overline{R(x)}\right)\left(R(y_i)-\overline{R(y)}\right)}{\sqrt{\sum_{i=1}^{n}\left(R(x_i)-\overline{R(x)}\right)^2\cdot\sum_{i=1}^{n}\left(R(y_i)-\overline{R(y)}\right)^2}} \\ &= 1-\frac{6\sum_{i-1}^{n}\left(R(x_i)-R(y_i)\right)^2}{n\left(n^2-1\right)}.\end{aligned} \tag{10}$$

*2) Feature Importance Analysis Based on Random Forest Regression*

Feature correlation analysis can be used to assess linear and ordinal consistent correlations but cannot identify other types of correlations. To address this limitation, this work proposes a feature importance analysis method to evaluate the impacts of different scenario features.

The decision tree is a non-parametric supervised learning algorithm that can be used in solving both classification and regression problems. It is a hierarchical tree structure mainly composed of three types of nodes, root, internal, and leaf nodes. Ensemble learning uses multiple models to obtain accurate prediction results, and bootstrapping is one of its typical application techniques that refers to the process of randomly sampling of a sub-dataset through a given number of iterations and variables. Random forest regression combines ensemble learning with the decision tree framework, creating multiple decision trees from data; then, multiple outputs are averaged to obtain the final result, often achieving excellent performance for regression problems.

Random forest regression can be used to evaluate feature importance. Particularly, in this study, the extracted scenario features are regarded as independent variables, while the prediction error and epistemic uncertainty of the prediction model under the corresponding conditions are regarded as dependent variables, and a random forest regression model is constructed. Then, the contribution of each feature to the trees in the random forest is analyzed, as well as its importance to the performance of trajectory prediction. The variable importance measure is denoted as VIM, and it is assumed that there are $J$ features and $I$ decision trees. Then, $\mathrm{VIM}_j$ denotes the average change in node split impurity of the $j$th feature in all decision trees, and it is calculated by:

$$\mathrm{VIM}_j=\frac{\sum_{i=1}^{I}\mathrm{VIM}_j^{(i)}}{\sum_{j'=1}^{J}\sum_{i=1}^{I}\mathrm{VIM}_{j'}^{(i)}}, \tag{11}$$

where $\mathrm{VIM}_j^{(i)}$ represents the importance of the $j$th feature in the $i$th decision tree, and it can be obtained by calculating the difference of Gini indices of nodes before and after branching.

*D. Prediction across Different Intersection Datasets*

The cross-dataset analysis aims to analyze differences in scenarios between multiple datasets and the corresponding prediction algorithm performance disparity. In this study, the scenario type is limited to the intersection, and multiple intersection datasets involving various countries are studied. First, the scenario features are extracted to analyze distributional shifts between different intersection datasets. Next, comprehensive cross-validation experiments are performed to investigate the prediction algorithm performance in terms of distributional shifts fully. Particularly, $N$ intersection datasets are selected, and each of them is divided into training and test subsets. Then, for each of the training subsets, a trajectory prediction model is developed and trained using the training subset first and then evaluated on the corresponding test subset. Finally, $N^2$ sets of results are obtained.

During the analysis, the following factors are mainly studied:

1) Trajectory prediction performance degradation due to distributional shifts: Combined with the differences in statistical features between different datasets obtained earlier, we analyze the impact of changes in the traffic environment on the prediction algorithm.

2) Effect of the deep ensemble on prediction robustness across different datasets: The improvement in prediction accuracy and sensitivity of the estimated epistemic uncertainty to distributional shifts are analyzed;
3) Availability and complexity of different intersection datasets for trajectory prediction: By synthesizing multiple groups of results, the model performance is evaluated using different intersection datasets as a training set and the prediction challenge with different intersection datasets as a test set.

## IV. Experimental Setup

### A. Intersection Datasets

Focusing on the urban intersection scenario, multiple trajectory datasets were used for evaluation and analysis, involving different periods, weather, countries and regions, and many TP types.

*1) SinD* [55]: The SinD dataset is a typical drone dataset collected from a signalized intersection in Tianjin, China. These data were recorded from a static bird's eye view at a sampling frequency of 10 Hz. This dataset contains about 420 minutes of traffic recordings, including over 13,000 TPs with seven types, including cars, trucks, buses, pedestrians, tricycles, bikes, and motorcycles;

*2) INTERACTION* [29]: The INTERACTION dataset contains 12 subsets covering merging, roundabout, and intersection scenarios, of which five intersection subsets are used in this study: USA_Intersection_EP1 (EP1), USA_Intersection_EP2 (EP2), USA_Intersection_MA (MA), USA_Intersection_GL (GL) and TC_Intersection_VA (VA). They contain about 493 minutes of recordings in total. The first four relate to unsignalized intersections in the US, mainly involving trajectories of vehicles, pedestrians, and bicycles recorded by drones. The VA denotes a signalized intersection in Bulgaria, which involves trajectories of cars, buses, trucks, motorcycles, and bicycles recorded by traffic cameras.

### B. Prediction Error Metrics

The prediction error is a preferred metric for quantifying the performance of prediction models. Following [2, 8, 25], the proposed model was evaluated using two error metrics:

*1) Average Displacement Error (ADE):* This is the mean square error of all predicted points of a trajectory compared to the ground truth;

*2) Final Displacement Error (FDE):* This is the distance between the predicted final destination and the true final destination at $t_f$ .

### C. Predictive Uncertainty Evaluation

As stated previously, the APE and FPE reflect the epistemic uncertainty of a prediction model, which can indicate situations where prediction models are performing poorly. Therefore, referring to [56, 57], this study uses the error-retention curves to evaluate the ability of the extracted epistemic uncertainty to detect prediction errors. The curves depict the error over a dataset as a model's predictions are replaced by ground-truth labels in order of decreasing uncertainty. The abscissa value of a point represents the proportion of the retained true error (i.e., retention fraction), while the ordinate value represents the comprehensive error under this proportion. Similarly, the optimal and random curves are obtained by replacing the predictions in order of decreasing error and random order, respectively. The area under the retention curves (AUC) is an evaluation metric of both the robustness of prediction models and the quality of uncertainty estimation, and an efficient uncertainty estimation is considered to achieve a low AUC.

### D. Implementation Details

For achieving fair evaluation and transferability, the datasets were standardized to use up to 3 s of the previous data and 3 s of the future data. The trajectory data were resampled to 2 Hz. The single GRIP++ model in the ensemble models was implemented in PyTorch, and its implementation details, including input preprocessing, graph convolution, and trajectory prediction model, mainly refer to the settings in [5]. In the implementation of MC dropout and deep ensemble, the value of $k$ was set to five, which was the result of a trade-off between uncertainty estimation quality and computational cost [45]. In addition, during the training process of the MC dropout-model, a regularization term was added to the loss to improve its ability of uncertainty estimation [39], where the regularization coefficient was set to 0.0001, and the dropout rate was set to 0.5

In addition, the original dataset was further processed to obtain the labels for the subsequent analysis. The locations of TAs were classified by combining their raw coordinate data with the map in the Lanelet2 format [58].

## V. Results and Discussion

### A. *Evaluation of Trajectory Prediction* and *Epistemic Uncertainty Estimation*

The training and test performances of the proposed model obtained by following the train-test process in the same intersection dataset are presented in TABLE I. Although the MC dropout can be used to estimate epistemic uncertainty, it increases the prediction error, which might be due to the modifications in the original loss function. In contrast, the deep ensemble-based method improves trajectory prediction accuracy while estimating epistemic uncertainty. As shown in TABLE I, the error obtained by the deep ensemble-based method was lower than that of the single models. Thus, by integrating the results of multiple models, deep ensemble could effectively avoid the prediction performance degradation caused by the deviation of a single model, which is conducive to improving the prediction algorithm robustness. The evaluation results of the estimated epistemic uncertainty are shown in Fig. 5. Compared with the MC dropout, the deep ensemble had obvious advantages in improving the model prediction accuracy and uncertainty estimation. Therefore, in the subsequent analysis, the epistemic uncertainty estimation framework based on the deep ensemble was adopted.

As shown in Fig. 5, both uncertainty quantification metrics (i.e., APE and FPE) could accurately reflect the prediction model error (i.e., ADE or FDE), and they showed a high degree of consistency. By comparing the left and right sides of Fig. 5, it can be concluded that although the prediction error on the test sets was slightly increased compared to that on the training set, there was a small difference in the epistemic uncertainty

TABLE I
TRAJECTORY PREDICTION ERROR COMPARISON[1]

| Dataset | $ADE_1/FDE_1$ | $ADE_1/FDE_1$ | $ADE_1/FDE_1$ | $ADE_1/FDE_1$ | $ADE_1/FDE_1$ | $ADE_{dropout}/FDE_{dropout}$ | ADE/FDE |
|---|---|---|---|---|---|---|---|
| **SinD** | 0.405/0.865 | 0.404/0.865 | 0.402/0.86 | 0.403/0.861 | 0.408/0.872 | 0.477/1.021 | **0.389/0.832** |
| **VA** | 0.652/1.401 | 0.641/1.378 | 0.655/1.428 | 0.657/1.421 | 0.654/1.402 | 0.651/1.327 | **0.615/1.327** |
| **EP0** | 0.811/1.822 | 0.815/1.844 | 0.803/1.809 | 0.809/1.814 | 0.822/1.850 | 1.036/2.351 | **0.745/1.678** |
| **EP1** | 0.881/1.982 | 0.842/1.894 | 0.816/1.826 | 0.853/1.915 | 0.857/1.930 | 1.147/2.590 | **0.775/1.743** |
| **MA** | 0.878/2.026 | 0.857/1.979 | 0.866//1.999 | 0.867/2.000 | 0.870/2.008 | 1.083/2.546 | **0.811//1.879** |
| **GL** | 0.619/1.448 | 0.627/1.464 | 0.622/1.452 | 0.625/1.458 | 0.627/1.466 | 0.709/1.646 | **0.591/1.386** |

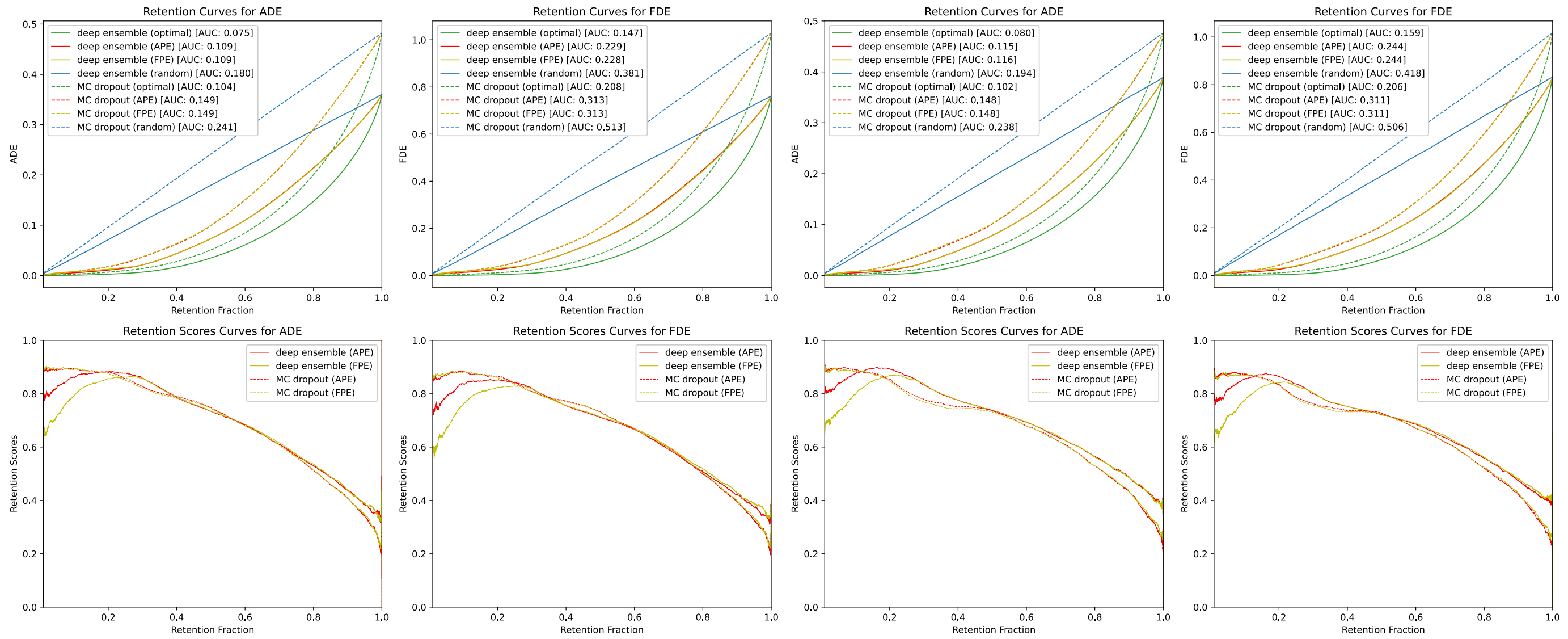

(a) ADE for training set (b) FDE for training set (c) APE for test set (d) FPE for test set

**Fig. 5.** The ADE/FDE-retention curves (top) and retention scores curves (bottom) on the training and test sets of the SinD dataset. The optimal curve (solid green line) was obtained by replacing the model's predictions with the ground-truth labels in order of decreasing error. Similarly, the random curve (blue dotted line) was obtained by replacing the model's predictions with the ground-truth labels in random order. The red and yellow solid lines correspond to the results captured in order of decreasing APE and decreasing FPE, respectively. The retention scores corresponding to each retention fraction can be calculated by: $(error_{random} - error_{uncertainty})/(error_{random} - error_{optimal})$.

estimation performance. The results indicate that the deep ensemble-based method had good generalization ability

In the second row in Fig. 5, a consistent trend where the retention scores first increase and then decrease with the retention fraction can be observed.

## *B. Scenario Features Analysis Results*

### *1) Feature Correlation Comparison*

In general, it has been considered that scenario complexity increases with the values of the extracted TA's kinematic features and the features of the TA's surrounding TPs. Therefore, the correlation between the model performance and these two types of features was calculated to analyze the influence of scenario complexity on trajectory prediction performance. Based on the model trained on the SinD training set, the feature correlation analysis experiment was performed on the SinD test set, where the prediction performance was represented by prediction error and epistemic uncertainty. For the features of the TA's surrounding TPs, four groups of distances of $x = 10, 20, 30, 50$ were studied. The analysis results are presented in Fig. 6, and based on them, the following conclusions can be drawn:

- The distribution trends of environmental feature correlations corresponding to the two types of prediction errors (ADE and FDE) were highly consistent, as well as the distribution trends of environmental feature correlations corresponding to the two types of epistemic uncertainty (APE and FPE) estimates;
- The comparison of a TA's kinematic features with the features of its surrounding TPs shows that the former had a strong positive correlation with the error and epistemic uncertainty of the prediction model, while the latter had weak correlations with the error and epistemic uncertainty ($-0.2 < \rho < 0.2$).

1 $ADE_k/FDE_k$ represents the prediction error of model $k$ among the ensemble models, $ADE_{dropout}/FDE_{dropout}$ is the prediction error of the mc dropout-based method, and ADE/FDE denotes the prediction error of the deep ensemble-based method.

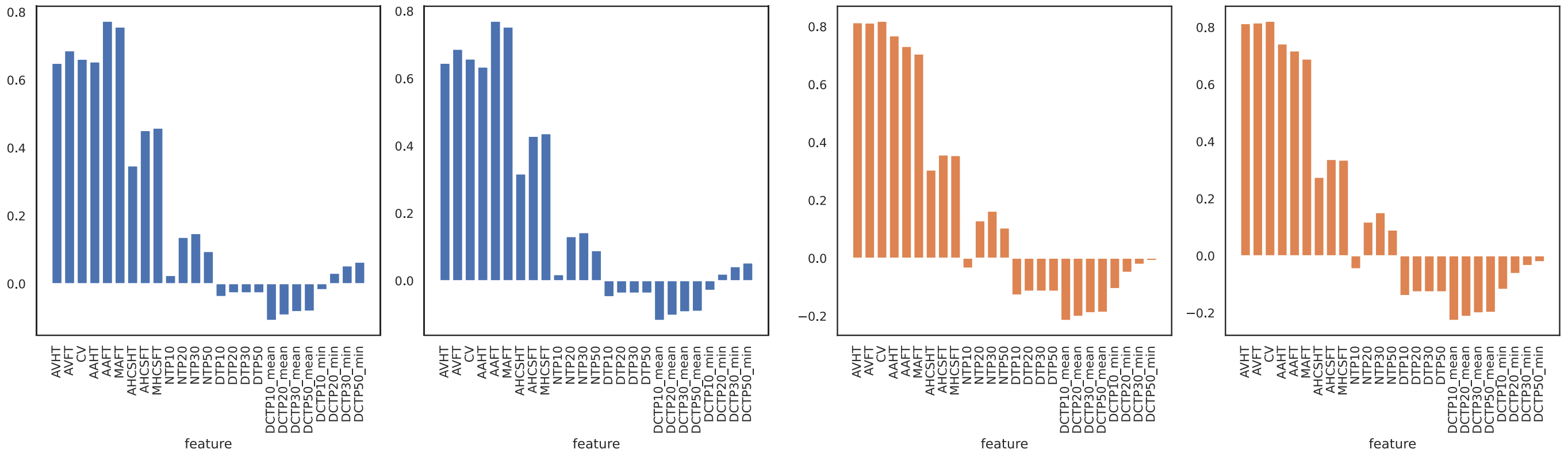

(a) $\rho$ for ADE (b) $\rho$ for FDE (a) $\rho$ for APE (b) $\rho$ for FPE

**Fig. 6.** Comparison of the correlation between prediction model performance and scenario features

- The comparison of the correlation between different kinematic features and the prediction error shows that:
  - The sorting order in terms of the correlation was: acceleration-related features > velocity-related features > heading change speed-related features. This order indicates that the mutation of a TA's speed had a relatively large impact on the prediction error, while the change in the TA's movement direction had a relatively low impact;
  - The sorting order in terms of the correlation was: features related to future trajectories > features related to historical trajectories. This shows that the proposed trajectory prediction model had low adaptability to the speed and position mutations occurring at some certain points in the future.
- The correlation between different kinematic sub-features of a TA and the predictive uncertainty showed that the epistemic uncertainty was highly sensitive to the velocity and acceleration features; namely, when a TA was driving at high speed or had a speed mutation, the model tended to show lower confidence in the predictions.

*2) Feature Importance Comparison*

As mentioned above, the feature correlation analysis only shows whether the relationship between two variables conforms to the order consistency. Therefore, the feature importance analysis experiment was performed based on the random forest regression to explore whether there were other types of correlation between the above scenario features and the prediction model. The datasets and training settings used for the prediction algorithm were the same as in the feature correlation analysis. The grid-based search was employed to obtain optimal random forest regression model, then the feature importance analysis was performed. The results are presented in Fig. 7.

As shown in Fig. 7, the distribution trend of feature importance was basically consistent with the feature correlation. For instance, the features obtained from the surrounding TPs had little effect on the error and uncertainty of the prediction model uniformly. In contrast, the kinematic features of a TA had a stronger influence, where velocity- and acceleration-related features had higher importance. In random forest regression for the ADE, the AAFT had the highest importance,

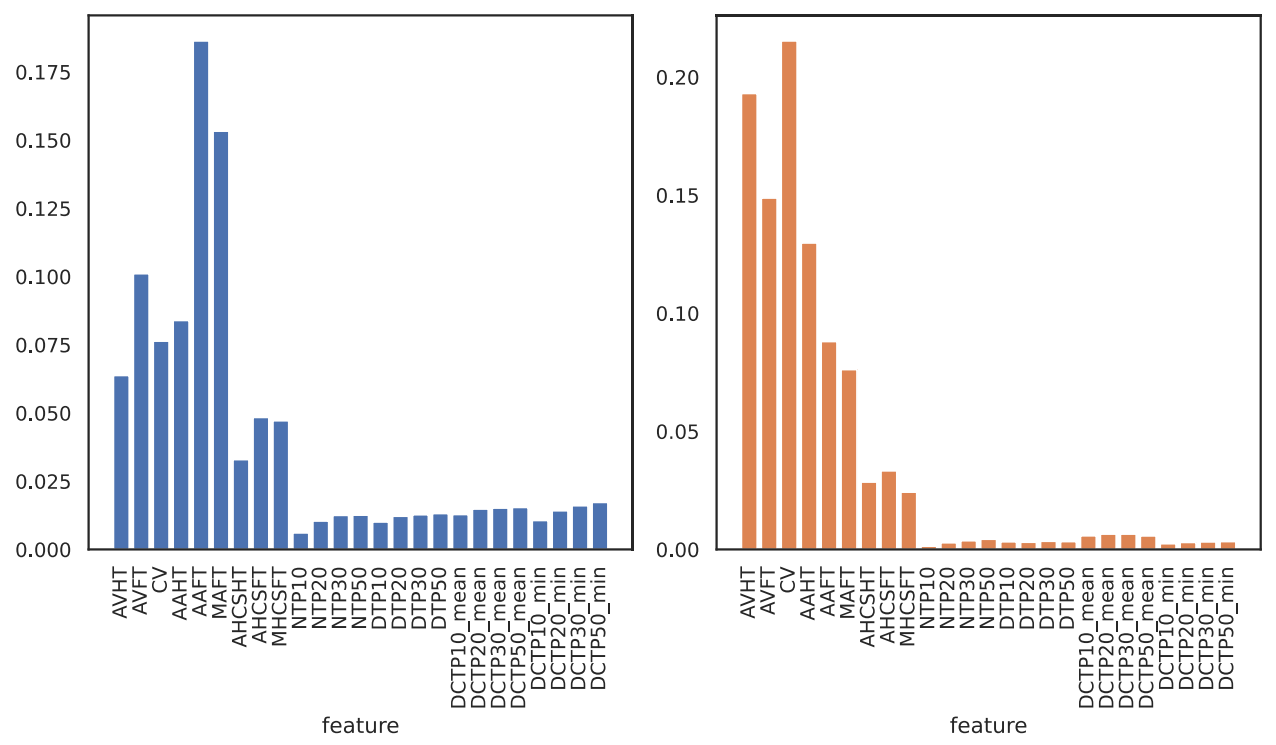

(a) feature importance for ADE (b) feature importance for APE

**Fig. 7.** Comparison of feature importance based on the random forest regression

while in the random forest regression for the APE, the CV had the strongest influence.

*3) Other Scenario Features Analysis*

In addition to the above-mentioned features, the impacts of several other environmental features on the prediction algorithm performance were explored, including the type, behavior patterns, compliance with traffic rules, and location of a TA. Without loss of generality, the ADE was adopted as an error metric, and the APE was used for epistemic uncertainty quantification. The prediction model was trained on the SinD training set and analyzed on the SinD test set

The results indicated obvious differences in the prediction error distribution between different TAs, as shown in Fig. 8. Although the movement of pedestrians had high degrees of freedom and randomness, their speed and acceleration were generally low, so the corresponding prediction error was small. Meanwhile, the epistemic uncertainty distribution for different types of TAs showed high similarity with that of the prediction error.

The results of the trajectory prediction error and epistemic uncertainty of a vehicle under different behavior patterns are presented in Fig. 9, where it can be seen that the trajectory prediction performance tended to exhibit larger errors when the TA was turning left and right compared to the going-straight behavior. The error in the right-turn scenario was relatively large, and the reasons may be as follows. First, the right-turn trajectory had a large curvature, and second, the vehicle was less affected by traffic lights and other TPs when turning right, compared to turning left and going straight, resulting in a higher driving speed. In addition, although the U-Turn represents a typical corner case, the results showed that the overall error in this pattern was small, which may be related to the generally low speed during the U-turning process. Furthermore, the epistemic uncertainty distributions under different behavioral patterns were relatively consistent with the error.

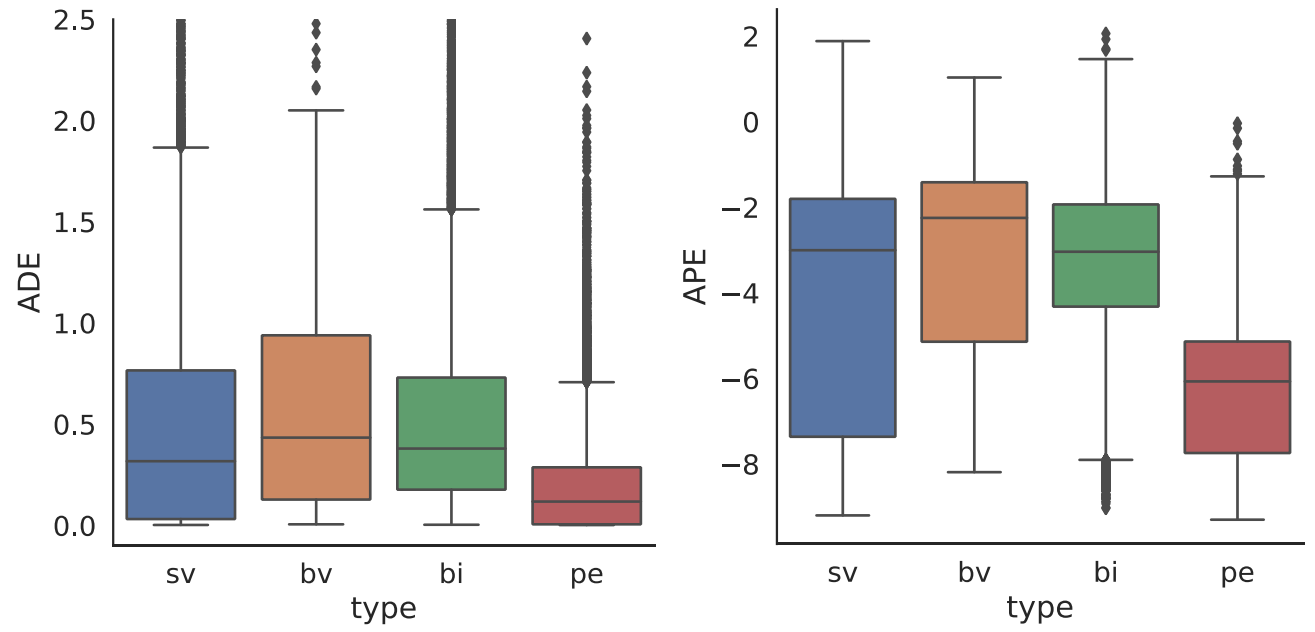

**Fig. 8.** The results of the trajectory prediction error and epistemic uncertainty of different TA types; sv: small vehicle, bv: large vehicle; bi: motorcyclist or bicyclist; pe: pedestrian.

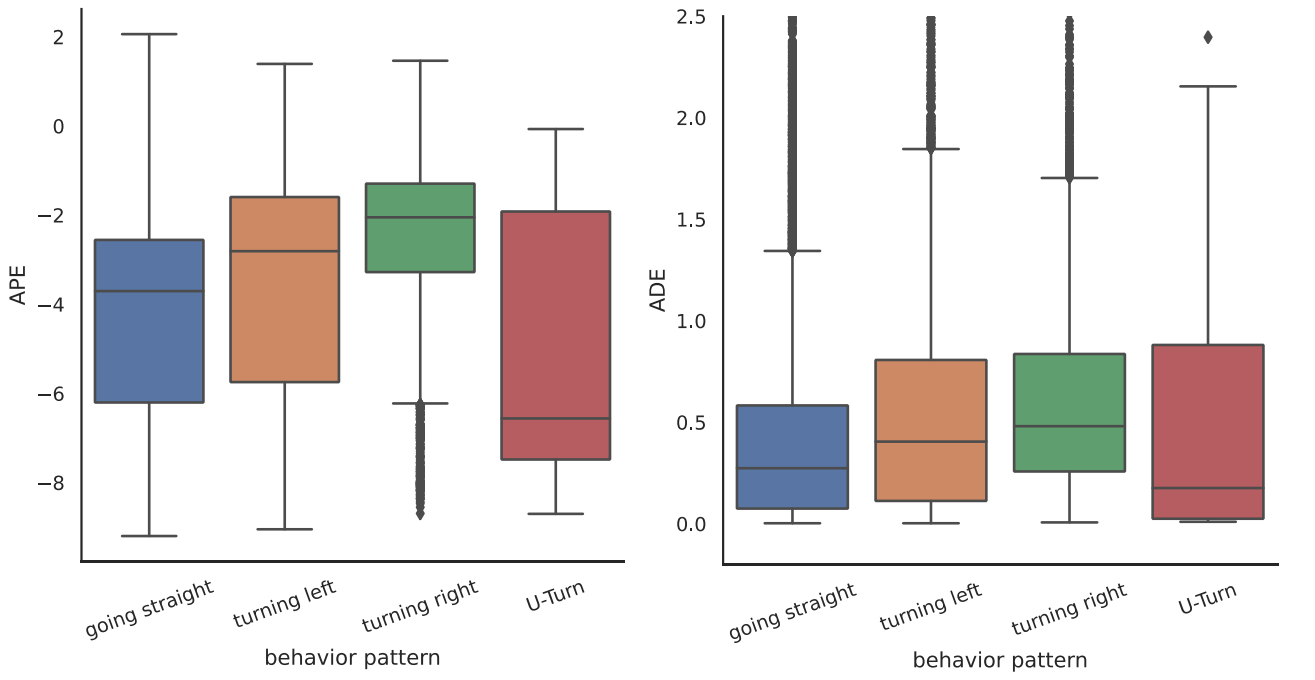

**Fig. 9.** The results of the trajectory prediction error and epistemic uncertainty under different behavioral patterns.

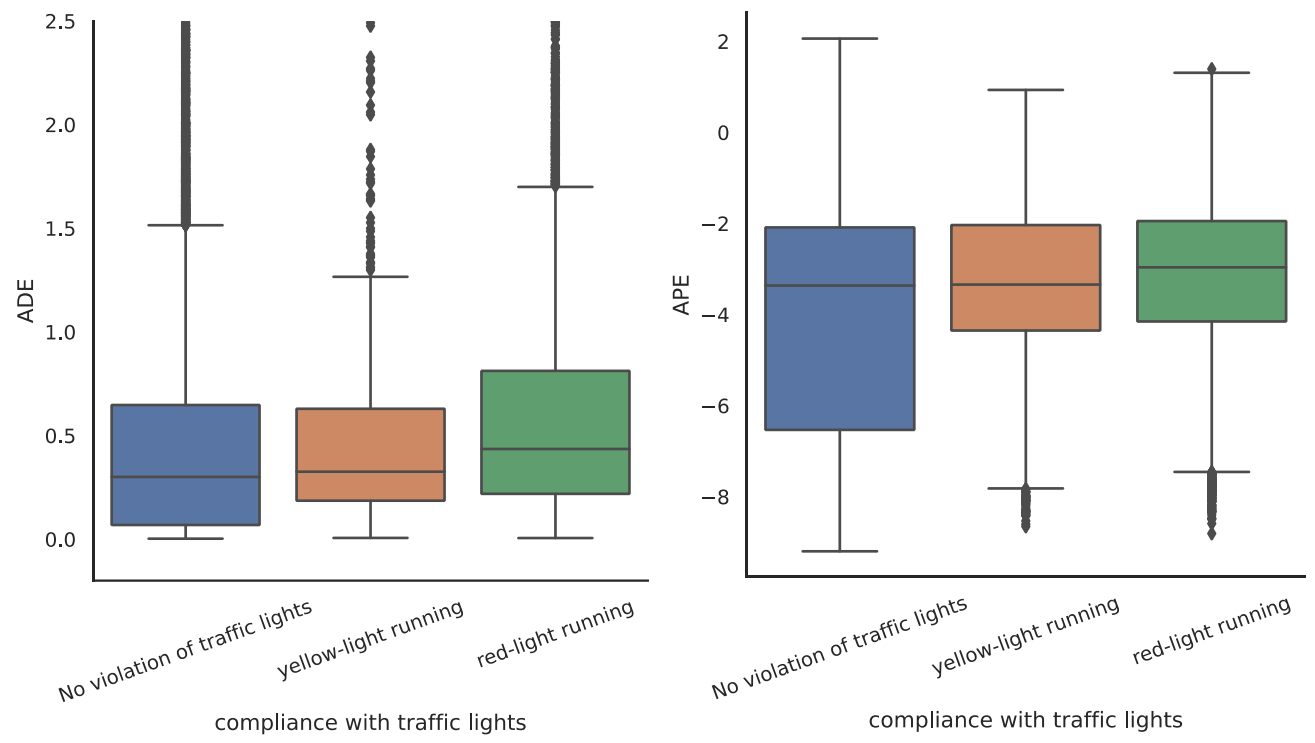

**Fig. 10.** The results of the trajectory prediction error and epistemic uncertainty under different traffic rule compliance.

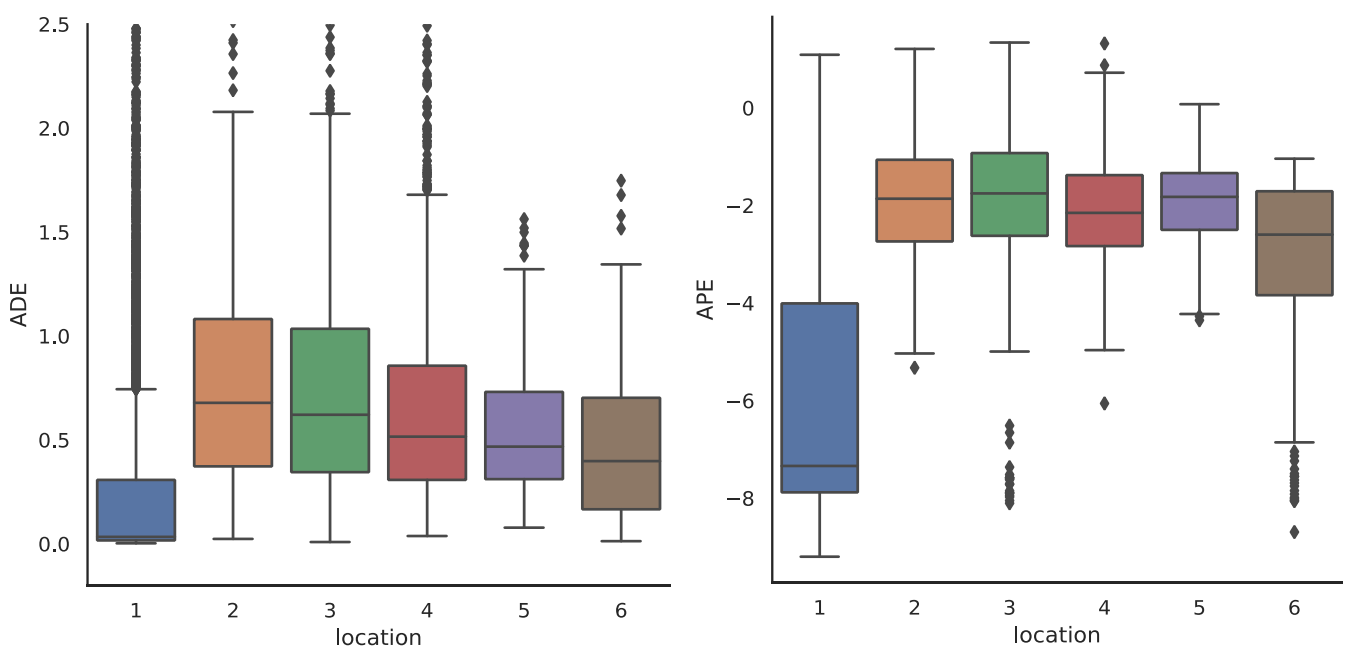

**Fig. 11.** The results of the trajectory prediction error and epistemic uncertainty at different locations; 1: ex-entering the intersection; 2: in the gap; 3: in the first crosswalk; 4: inside the intersection; 5: in the last crosswalk; and 6: exiting the intersection.

The relationship between the vehicles' compliance with traffic lights and the trajectory prediction performance is presented in Fig. 10, where it can be seen that the prediction error was larger when the vehicle ran red or yellow light than when there was no violation of traffic lights, and simultaneously the proposed model could output higher epistemic uncertainty.

The impact of a TA's location on the trajectory prediction performance is presented in Fig. 11. When the vehicle was in the gap area or first crosswalk before entering the intersection, there were many possible strategic options, referring to whether to enter the intersection and how to pass through the intersection, which increased the prediction complexity, and further increased the prediction error and epistemic uncertainty. When the vehicle was inside the intersection, the prediction model exhibited considerable error and uncertainty due to a large number of interactions with other TPs and higher freedom of movement. In contrast, the predominant behavior of the vehicles before entering and after exiting the intersection was to follow the lane, so these stages had lower prediction error and uncertainty than the others.

### *C. Prediction evaluation across Different Intersection Datasets*

Different intersection datasets were collected at different times and locations, and the corresponding environmental conditions might be relatively different, resulting in shifts in data distribution. The distributions of velocity, acceleration, heading, and HCS of objects in six intersection datasets are presented in Fig. 12, where it can be seen that there were obvious differences in the trajectory features between the datasets. For instance, the velocity and acceleration of a portion of trajectories in the SinD dataset were concentrated around zero. One of the main reasons was the stopping of vehicles and pedestrians while waiting for the green light. Furthermore, the velocity in the SinD dataset exhibited a distinct multimodal distribution, which could be related to the multiple movement patterns caused by various TPs in the dataset. In contrast, the velocity and acceleration in the GL, MA, and VA datasets tended to be higher, reflecting more aggressive motions in these datasets. In addition, distributional shifts could also be observed by comparing the distribution of heading and its speed of change in different datasets.

The aforementioned differences in data distribution between datasets may bring application challenges of the trajectory prediction algorithms in real-world environments. Therefore, experiments were performed on multiple intersection datasets. Specifically, based on the six intersection datasets mentioned above, a set of deep ensemble-based prediction models were trained on each dataset and evaluated on all six datasets. As shown in Fig. 13, distributional shifts in the real-traffic environments had a strong impact on trajectory prediction performance. Even for the same type of scenario, for the model trained on one intersection dataset, it was difficult to generalize well to other intersection datasets directly. For instance, on the test subset of the SinD dataset, the model that achieved the best accuracy ($ADE_1$/ADE: 0.405/0.389) was trained on the training subset of the SinD dataset; in contrast, the prediction error ($ADE_1$/ADE) of the model trained on the other intersection training subsets increased by 94.9%/8.61% - 265.0%/221.0%.

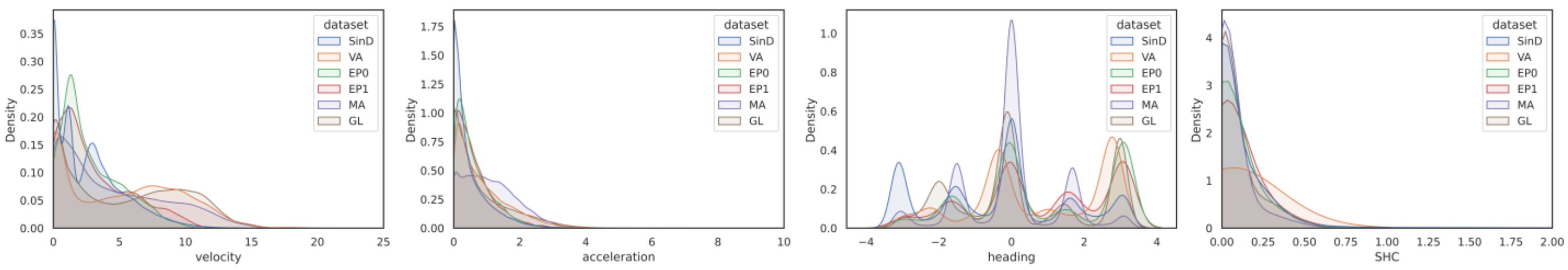

**Fig. 12.** Distribution comparison of different intersection datasets.

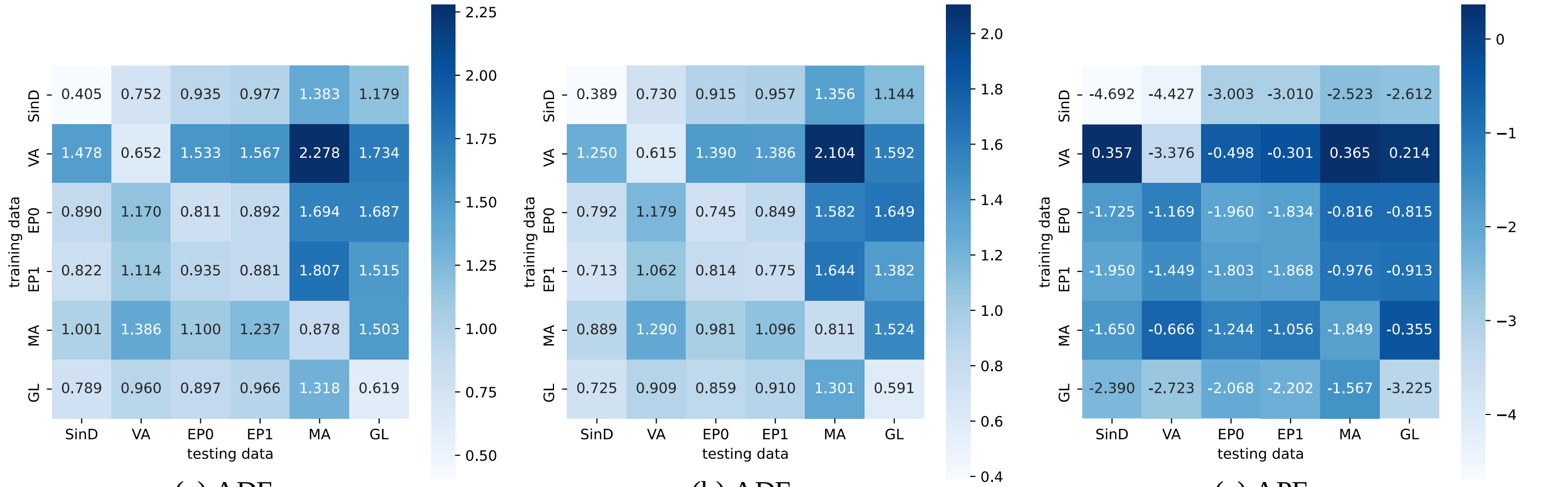

(a) $ADE_1$ (b) ADE (c) APE

**Fig. 13.** The cross-dataset error and uncertainty matrix: (a) $ADE_1$ is the error obtained by evaluating one of the ensemble models; (b) ADE is the prediction error of the deep ensemble-based model; (c) APE also relates to the deep ensemble-based model. The *i*th row and *j*th column of each matrix represent the results obtained by evaluating the model on the test subset at intersection *j* after training it on the training subset at intersection *i*.

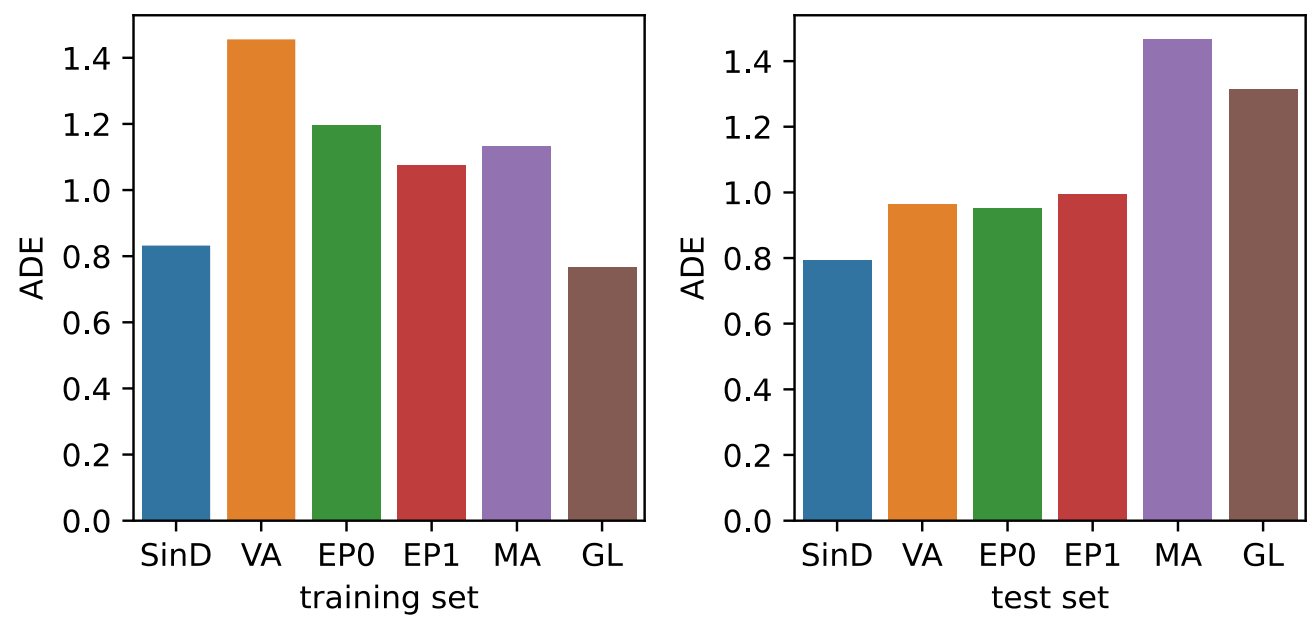

**Fig. 14.** Comprehensive performances of the prediction algorithms on different datasets. (a) Comparison of prediction errors of all trained models on different test subsets; (b) results of evaluating models trained on different training subsets on all test subsets.

Comparing the single prediction model error (Fig. 13(a)) with the error of the deep ensemble-based model (Fig. 13(b)), it could be concluded that the deep ensemble-based approach performed systematically better than a single model, achieving significant improvements in many cases.

As presented in Fig. 13(c), the extracted epistemic uncertainty could indicate distributional shifts between different datasets. Particularly, the proposed model could output higher epistemic uncertainty when the error increased due to changes in the test scenario.

In Fig. 14, the comprehensive performances of the trajectory prediction algorithm on specific single training or test set are presented. As shown in Fig. 14(a), the models trained on the training subsets of the SinD and GL datasets had better generalization ability than the other models. This could be because of a larger amount of data and more diverse motion patterns of these two datasets compared to the other datasets. Conversely, the overall error of the model based on the training subset of VA was the highest, and the main reasons were as follows. First, this dataset contained less data than the other datasets. Second, this dataset was constructed by data collected by roadside equipment, which might introduce more noise than drone-based data collection. The comprehensive performances of the prediction models on different intersection test subsets are presented in Fig. 14(b). The GL and MA with higher velocity and acceleration features showed greater prediction difficulty, and the overall trajectory prediction error on the SinD test subset was the smallest among all datasets.

### *D. Qualitative Results*

The prediction results of the proposed framework in several scenarios on different intersection datasets are presented in Fig. 15, where each row corresponds to one intersection. In Fig. 15, the first column shows cases where the prediction error was small and epistemic uncertainty was low. These cases generally appeared when the TA exhibited obvious future intentions and moved relatively smoothly. The second and third columns show scenarios with large trajectory prediction errors, which were generally accompanied by higher estimates of epistemic uncertainty. This situation may occur when the TA was about to enter the intersection where it was difficult to determine its future behavior pattern, or its future motion showed a large pattern or speed change compared to the historical trajectory.

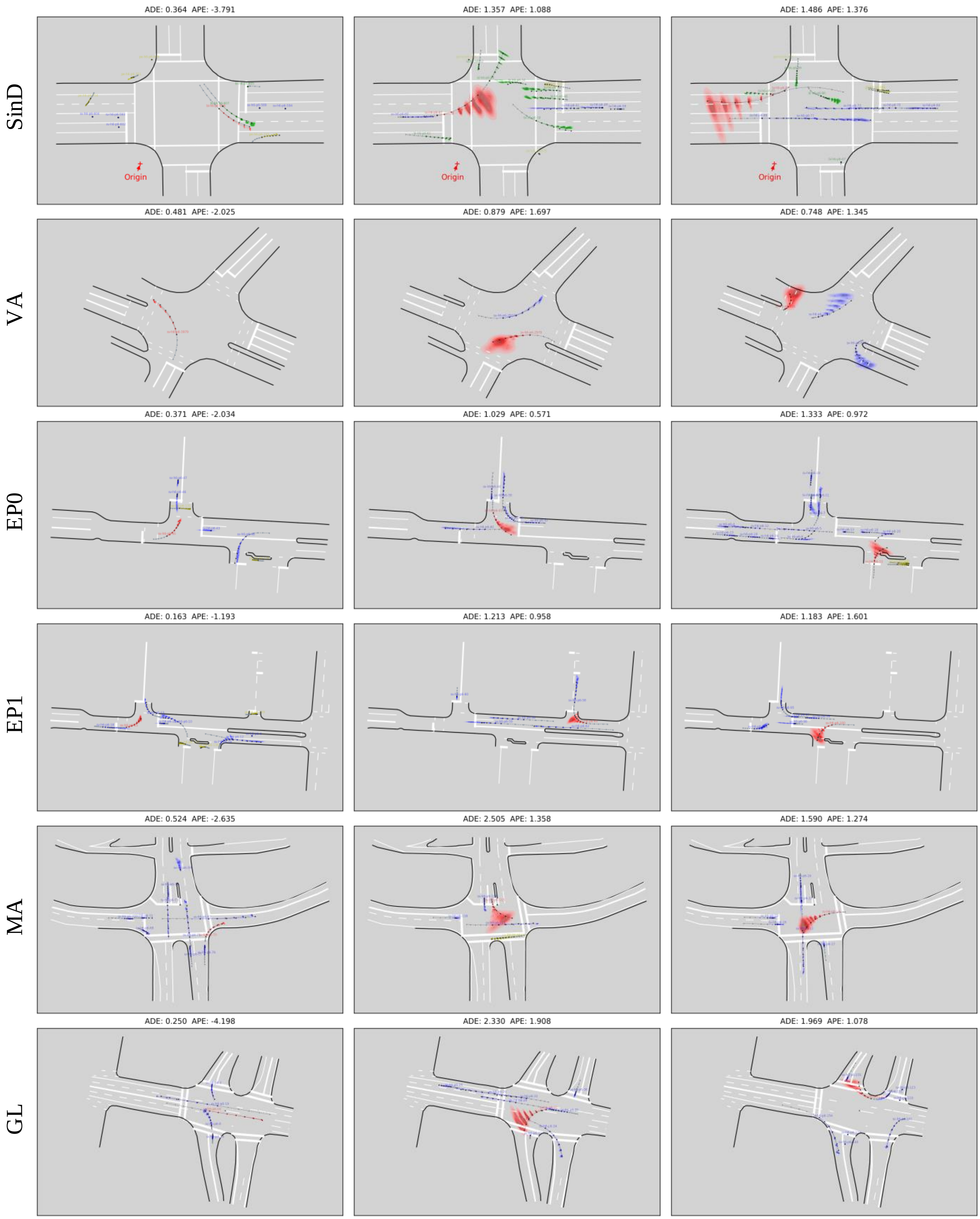

**Fig. 15.** Qualitative results of trajectory prediction and uncertainty estimation on different datasets (corresponding to different rows). The gray thick solid line represents the historical trajectory, and the black thin solid line represents the true future trajectory. Different colors are used to denote predictions for different types of TAs: blue - small vehicle; magenta - large vehicle; yellow – pedestrian; green - motorcycle or cyclist. In addition, the focused object in each subgraph is highlighted in red, and the current prediction error and uncertainty estimation results of the object are presented above the subgraph.

## VI. Conclusion

In this paper, a trajectory prediction framework that integrates the epistemic uncertainty estimation function is proposed, and the effects of the traffic environment and its changes on the prediction algorithm performance are studied. A few typical scenario features are considered, and their influences on the prediction performance are examined by the feature correlation and importance analyses. Further, the distributional shifts between different intersection datasets and the resulting performance degradation of the prediction model are analyzed. Based on the obtained results, the following conclusions are drawn:

(1) The extracted epistemic uncertainty is valuable for representing the model's confidence in its current predictions accurately, which has the potential to be used to identify unknown scenarios where the model may be underpowered. Compared with the MC dropout-based method, the deep ensemble-based method performs significantly better in estimating epistemic uncertainty and improving the trajectory prediction robustness;

(2) Regarding the influence of different scenario features on the trajectory prediction performance, the feature correlation and importance analyses show similar results. Namely, there is a positive correlation between the kinematics of a TA and the prediction model performance. Higher velocity, acceleration, and speed of the heading change generally pose a greater challenge to the trajectory prediction process, while a prediction model tends to exhibit higher epistemic uncertainty. However, one interesting finding is that the features of surrounding TPs, which reflect the complexity of interactions in the scenario, show little impact on the proposed prediction model performance. In addition, the error and uncertainty of the prediction model vary with other abstract features, such as TA's type, behavior pattern, compliance with traffic rules, and location. The conducted analyses are helpful in locating the limitations in the prediction algorithms, thus providing guidance for the improvement of autonomous driving functions;

(3) For the intersection scenario, different datasets show distributional shifts due to differences in local driving habits, road structures, and national cultures, thus posing great challenges to the prediction algorithms. Fortunately, the proposed framework based on the deep ensemble is beneficial to improving the trajectory prediction robustness, and the extracted epistemic uncertainty can respond to the reduced confidence of the proposed model in a new environment. This improves the self-awareness ability of autonomous driving.

Although the used basic prediction algorithm has strong representativeness, it is still difficult to avoid the specificity of analysis conclusions completely. However, the proposed method is promising to analyze other prediction algorithms, subsequent work could consider combining more types of algorithms for systematic analysis. Moreover, this work focuses on the extraction and analysis of epistemic uncertainty, but the existing works in the field of multimodal trajectory forecasting could be considered in the follow-up research to distinguish the epistemic uncertainty from the aleatoric uncertainty better and explore their practical significance. Furthermore, the role of uncertainty estimation of a trajectory prediction model in an autonomous driving decision-making mechanism needs to be studied further to improve the robustness against the risks of insufficient functions.

## ACKNOWLEDGMENT

This work was supported in part by the National Science Foundation of China Project (Grant No. 52072215 and U1964203), and the National Key R&D Program of China under Grant NO. 2020YFB1600303.

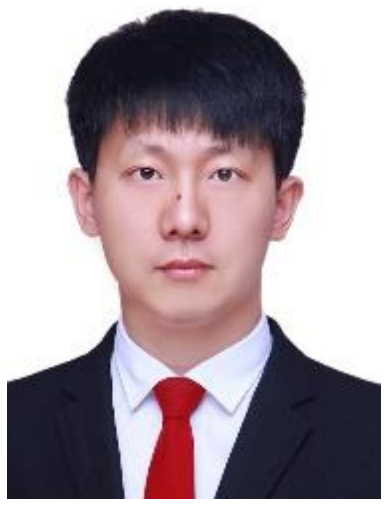

**Wenbo Shao** received his B.E. degree in vehicle engineering from Tsinghua University, Beijing, China, in 2019. He is currently working toward the Ph.D. degree in Mechanical Engineering at Tsinghua University. He is a member of Tsinghua Intelligent Vehicle Design And Safety Research Institute (IVDAS) and supervised by Professor Jun Li and Associate Research Professor Hong Wang. His research interests include safety of the intended functionality of autonomous driving, trajectory prediction, decision-making, uncertainty theory and applications.

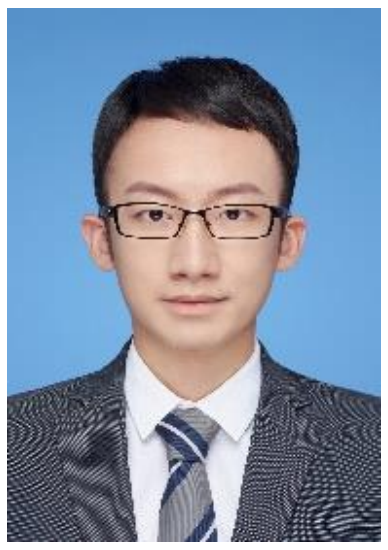

Yanchao Xu received B.E degree in vehicle engineering from Hainan University in China in 2020. He is currently pursuing the M.S. degree in Mechanical Engineering at Beijing Institute of Technology. He is also one of the visiting students at IVDAS since 2020. His research interest includes prediction, trajectory data mining, scenario parameterization for autonomous driving.

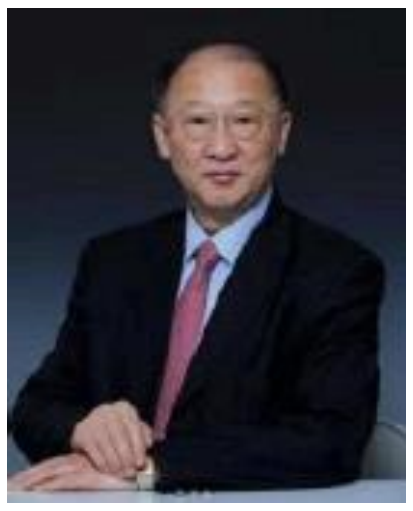

**Jun Li** received the Ph.D. degree in vehicle engineering from Jilin University, Changchun, Jilin, China, in 1989. He is currently an academician of the Chinese Academy of Engineering, a Professor at school of Vehicle and Mobility with Tsinghua University, president of the Society of Automotive Engineers of China, director of the expert committee of China Industry Innovation Alliance for the Intelligent and Connected Vehicles. His research interests include internal combustion engine, electric drive systems, electric vehicles, intelligent vehicles and connected vehicles.

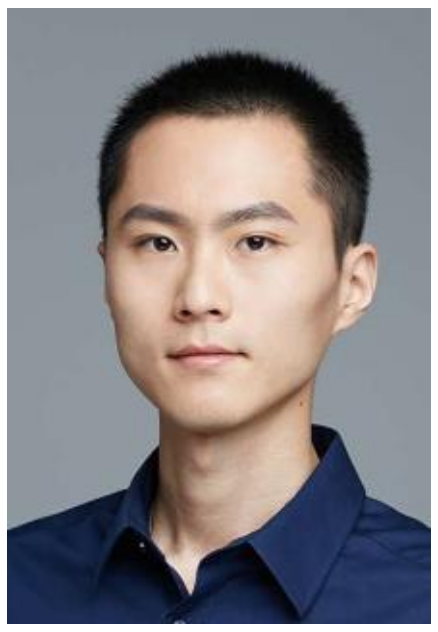

**Chen Lv** (Senior Member, IEEE) received the Ph.D. degree from the Department of Automotive Engineering, Tsinghua University, China, in 2016. From 2014 to 2015, he was a Joint Ph.D. Researcher at the EECS Department, University of California at Berkeley, Berkeley, CA, USA. From 2016 to 2018, he worked as a Research Fellow at the Advanced Vehicle Engineering Center, Cranfield University, U.K. He is currently an Assistant Professor with the School of Mechanical and Aerospace Engineering, and the Cluster Director of future mobility solutions at ERI@N, Nanyang Technological University, Singapore. His research interests include advanced vehicles and human–machine systems, where he has contributed over 100 articles and obtained 12 granted patents.

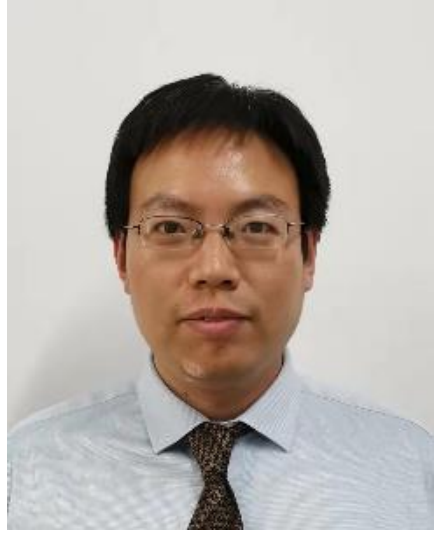

**Weida Wang** received the Ph.D. degree from Beihang University, Beijing, China, in 2009. He is currently a Professor with the School of Mechanical Engineering, Beijing Institute of Technology, Beijing. He is also the Director of the Research Institute of Special Vehicle, Beijing Institute of Technology. His current research interests include electric vehicle, automated vehicle motion planning and control, and electromechanical transmission control.

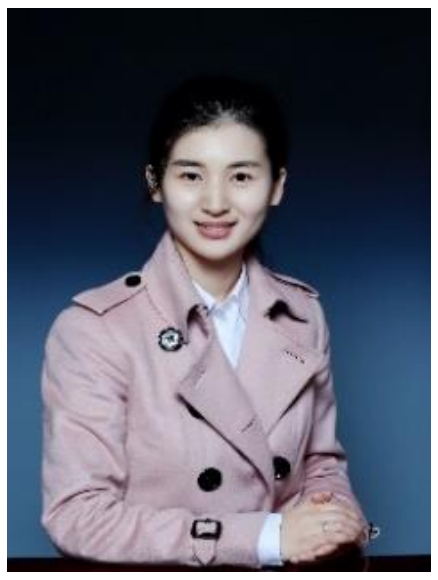

**Hong Wang** is Research Associate Professor at Tsinghua University. She received the Ph.D. degree in Beijing Institute of Technology, Beijing, China, in 2015. From the year 2015 to 2019, she was working as a Research Associate of Mechanical and Mechatronics Engineering with the University of Waterloo. Her research focuses on the safety of the on-board AI algorithm, the safe decision-making for intelligent vehicles, and the test and evaluation of SOTIF. She becomes the IEEE member since the year 2017. She has published over 60 papers on top international journals. Her domestic and foreign academic part-time includes the associate editor for IEEE Transactions on Vehicular Technology and Intelligent Vehicles Symposium, Guest Editor of Special Issues on Intelligent Safety of Automotive Innovation, Young Communication Expert of Engineering, lead Guest Editor of Special Issues on Intelligent Safety of IEEE Intelligent Transportation Systems Magazine.